\documentclass[letterpaper]{article} 
\usepackage{aaai2027}  
\usepackage[hyphens]{url}  
\usepackage{graphicx} 
\usepackage{natbib}  
\usepackage{caption} 
\usepackage{algorithm}
\usepackage{algpseudocode}
\usepackage{amsmath} 
\usepackage{multirow}
\usepackage{array}
\usepackage{newfloat}
\usepackage{listings}
\DeclareCaptionStyle{ruled}{labelfont=normalfont,labelsep=colon,strut=off} 
\floatstyle{ruled}
\newfloat{listing}{tb}{lst}{}
\floatname{listing}{Listing}

\usepackage{booktabs}

\title{TSS: Target-Side Sparsification for Speculative Decoding in Domain-Specific Large Language Models}
\author{
    Haibo Hu$^{1}$\quad Lianming Huang\textsuperscript{\rm 1}\thanks{Corresponding author:lmhuang8-c@my.cityu.edu.hk}\quad Qiao Li\textsuperscript{\rm 2}\quad  Nan Guan$^{1}$\quad  Chun
    Jason Xue$^{2}$ \\
    $^{1}$City University of Hong Kong\\
    $^{2}$Mohamed bin Zayed University of Artificial Intelligence
}
\affiliations{
}

\begin{document}
\nocopyright
\maketitle

\begin{abstract}
Speculative decoding accelerates large language model inference through collaboration between a lightweight draft model and a target verifier. Existing methods mainly improve the draft side, while the target model is typically kept dense and unchanged. We show that, under domain-specific inference, full-depth target verification is not always the optimal choice. Counter-intuitively, skipping selected target layers can reduce verification cost while simultaneously increasing draft acceptance and preserving, or even improving, downstream task performance. Based on this observation, we propose \textbf{TSS}, a target-side sparsification framework for speculative decoding. TSS employs an acceptance- and metric-aware breadth search to explore multi-layer skip configurations without imposing a fixed priority between the two objectives. The selected configurations are stored in a domain-to-configuration mapping and applied by a lightweight skip controller, allowing one complete target model to support multiple sparse verification paths without retraining or permanent parameter pruning. Experiments on Spec-Bench across multiple domains, model scales, and speculative decoding methods show consistent improvements in draft acceptance and downstream task performance. In Translation setting, TSS increases the average accept length from 2.70 to 4.53 ($+67.8\%$), improves BLEU from 0.131 to 0.237 ($+80.9\%$), and raises end-to-end throughput from 75.6 to 127.3 tokens/s, corresponding to a $1.68\times$ speedup.
\end{abstract}

\section{Introduction}
\begin{figure}[t]
  \centering
  \includegraphics[width=0.5\textwidth,trim=0cm 10.5cm 11cm 0cm, clip]{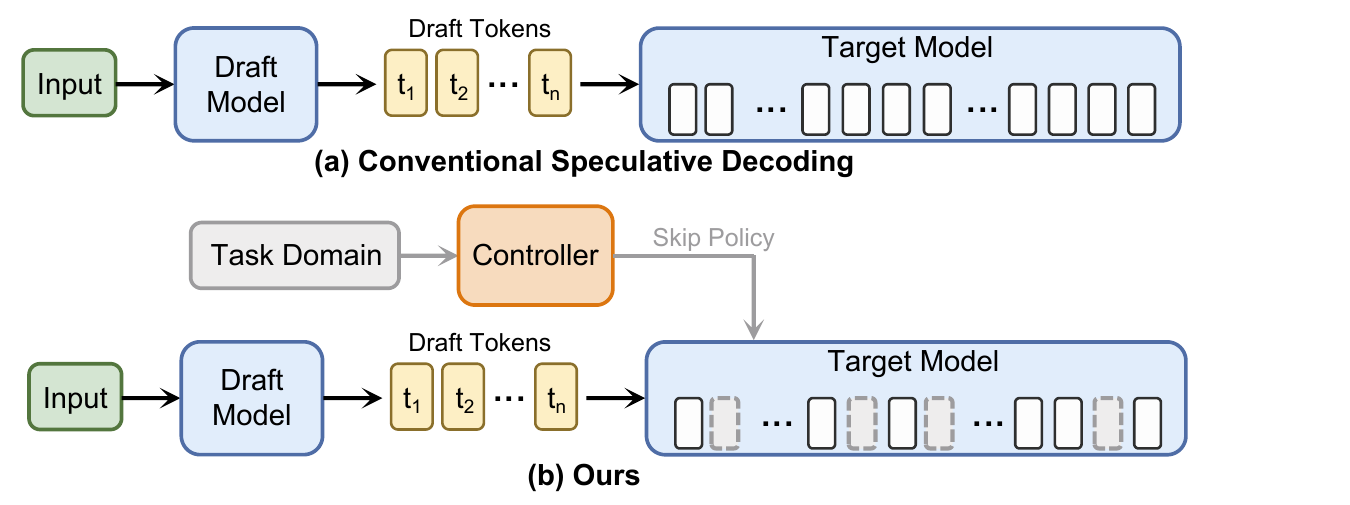}
  \vspace{-10pt}
\caption{
Comparison between conventional speculative decoding and TSS.
}
  \vspace{-10pt} 
  \label{fig:target}
\end{figure}

Large language models (LLMs) have become the foundation of modern AI applications, yet their deployment remains constrained by the cost of autoregressive decoding, where a Transformer decoder typically generates only one token per forward pass~\cite{pope2023efficiently,leviathan2023fast,chen2023accelerating}. This sequential dependency has motivated a broad range of acceleration methods, including draft-based speculative decoding~\cite{leviathan2023fast,chen2023accelerating}, sequence-to-sequence speculation~\cite{xia2023speculative}, big-little model collaboration~\cite{kim2023speculative}, tree-based verification~\cite{miao2024specinfer,spector2023accelerating}, multi-head candidate generation~\cite{cai2024medusa}, feature-level speculative sampling~\cite{li2024eagle}, and draft-free parallel decoding~\cite{fu2024lookahead,zhao2024lookahead}. As illustrated in Figure~\ref{fig:target}, these methods largely follow a draft-target pipeline in which candidate generation is optimized while the full target model remains densely activated during verification. Our work starts from a different observation: in domain-specific speculative decoding, full-depth target verification may over-process representations that are already sufficient for the downstream task. Prior studies describe this behavior as \emph{overthinking}, where correct intermediate predictions are later revised into incorrect outputs~\cite{kaya2019shallow,halawi2024overthinking}, while layer-wise decoding studies show that intermediate and final Transformer layers expose different predictive information~\cite{chuang2024dola}. Motivated by this observation, we introduce a domain-aware skip controller that selects a target-layer skip policy without modifying the draft model or candidate proposal process. During verification, the target follows a domain-specific sparse path and bypasses layers likely to induce unnecessary refinement. Empirically, this design not only reduces verification cost, but also increases draft acceptance while preserving or improving downstream task metrics.

However, finding an effective skip configuration for speculative decoding is challenging. First, conventional speculative decoding treats the dense target model as the reference distribution, and many compression or sparsification methods therefore rely on loss, perplexity, or distributional similarity to preserve the original model behavior. Once the target verifier is sparsified, however, its token distribution inevitably changes. For domain-specific generation, preserving the dense target distribution is not necessarily equivalent to preserving the downstream task metric, making loss-based layer selection a suboptimal objective. Second, speculative decoding introduces an additional system-level requirement: the sparse verifier must maintain or improve draft acceptance. If the accept length becomes too low, speculative decoding degenerates into ordinary autoregressive decoding with extra draft overhead. Therefore, target-side sparsification requires a new skip-layer search criterion that jointly considers downstream metrics and draft acceptance. 

Based on these observations and challenges, we propose \textbf{TSS}, a target-side block-layer sparsification framework for speculative decoding in domain-specific LLM inference. TSS searches for sparse target configurations offline on a calibration set, with downstream task metrics as the primary quality constraint and draft acceptance as a key speculative decoding objective. Since the skip pattern is selected offline, TSS can afford a broader search over multi-layer skip combinations rather than relying only on local single-layer scores. The final sparse verifier is chosen to optimize a dual objective: preserving or improving domain-level metrics while increasing accepted draft tokens and reducing target-side computation. Our contributions are summarized as follows.

\begin{itemize}
    \item \textbf{Target-side sparsification for speculative decoding.}
   To our knowledge, TSS is the first framework that treats the deployed target verification path itself—rather than a layer-skipped self-draft—as the domain-specific sparsification target. TSS directly skips selected target layers to reduce verification cost while preserving, and sometimes improving, downstream task performance.

    \item \textbf{Acceptance- and metric-aware skip selection.}
    We propose a skip-layer search algorithm tailored to speculative decoding. Instead of selecting layers according to language-modeling loss, perplexity, or distributional similarity, TSS uses downstream task metrics and draft acceptance as the core objectives.

\item \textbf{Efficient domain-aware skip control and broad validation.}
TSS searches skip configurations offline and applies them through a lightweight domain-aware controller. We evaluate TSS on Spec-Bench across multiple domains, model scales, and speculative decoding methods. The results show substantial throughput gains, higher accept length, and stable or improved task metrics using directly reproducible skip configurations.
\end{itemize}

\section{Related Work}

\subsection{Speculative Decoding and Draft-Side Optimization}

Speculative decoding accelerates autoregressive generation by using a lightweight draft model to propose candidate tokens and a stronger target model to verify them in parallel~\cite{leviathan2023fast,chen2023accelerating}. Most subsequent work focuses on improving the draft or proposal process. SpecDec trains an independent drafter~\cite{xia2023speculative}, BiLD uses big-little model collaboration~\cite{kim2023speculative}, and SpecInfer and staged speculative decoding organize candidates into trees or stages for parallel verification~\cite{miao2024specinfer,spector2023accelerating}. Medusa and EAGLE improve proposals through multiple decoding heads or feature prediction~\cite{cai2024medusa,li2024eagle}, while Lookahead decoding explores draft-free parallel generation~\cite{fu2024lookahead}.KNN-SSD further classifies incoming requests by domain and matches prompts to domain-specific skip configurations, demonstrating the feasibility of prompt-level domain identification~\cite{song2026knnssd}. Despite their differences, these methods generally keep the target model dense and fixed. Instead, TSS treats the target verifier itself as an optimization target.

\subsection{Layer Skipping and Sparse LLM Inference}

Another line of work reduces Transformer computation through layer dropping, early exiting, or direct layer removal. LayerDrop enables subnetworks of different depths~\cite{fan2020layerdrop}, while DeeBERT, FastBERT, and PABEE use intermediate confidence signals for early exit~\cite{xin2020deebert,liu2020fastbert,zhou2020bert}. SkipBERT skips shallow layers using approximate representations~\cite{wang2022skipbert}, CALM and RAEE dynamically adjusts the computation during generation~\cite{schuster2022confident, huang2024raee}. Recent LLM-oriented methods, including LayerSkip, FREE, and ShortGPT, further suggest that full-depth generation may involve unnecessary late-layer computation and overthinking~\cite{elhoushi2024layerskip,bae2023free,men2025shortgpt}. These methods mainly optimize perplexity, confidence, representation similarity, or standalone accuracy. In contrast, TSS mitigates domain-conditioned overthinking by selecting target layers according to both speculative acceptance and downstream task metrics.

\section{Motivation}
\begin{figure}[t]
  \centering
  \includegraphics[width=0.5\textwidth,trim=0cm 0.5cm 18cm 0cm, clip]{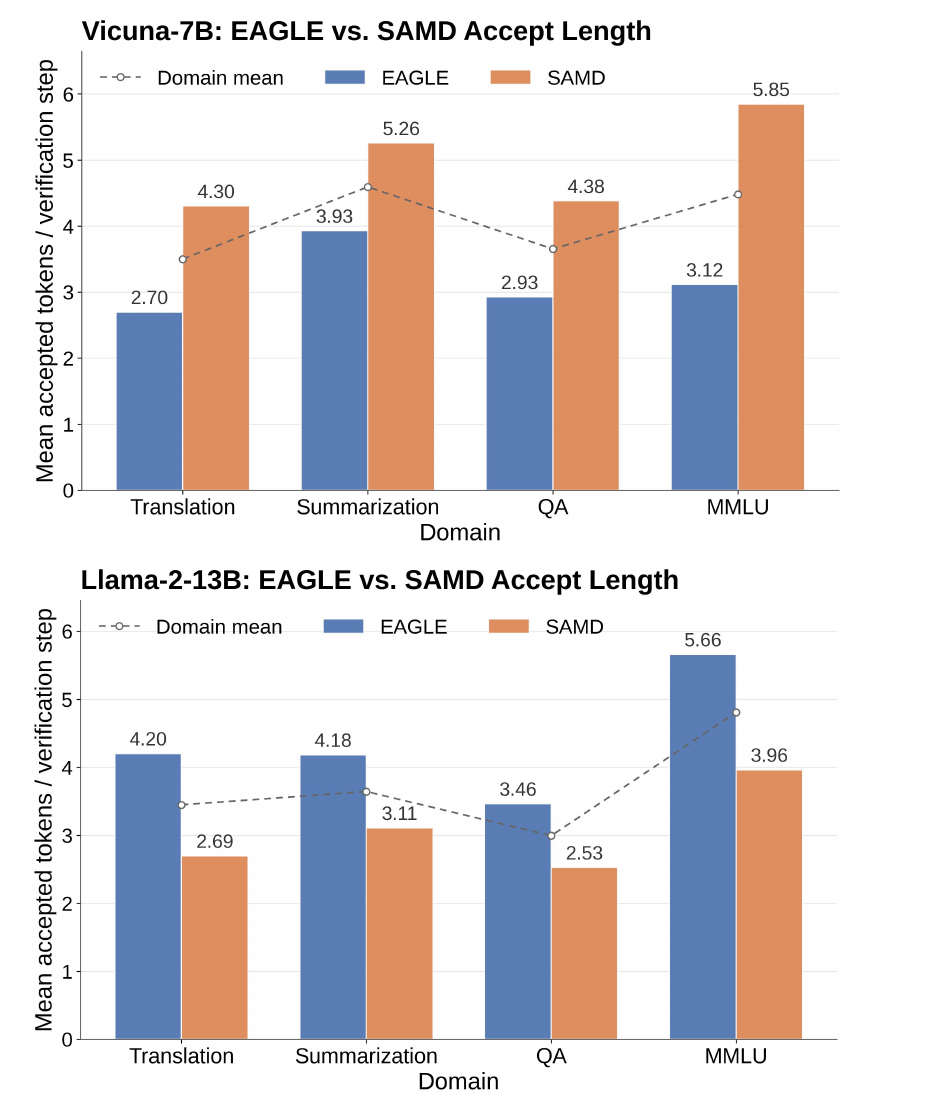}
  \vspace{-10pt}
\caption{
Mean accepted tokens per verification step for EAGLE and SAMD on Vicuna-7B and Llama-2-13B across four domains. 
}
  \vspace{-10pt} 
\label{fig:domain_accept_length}
\end{figure}
\subsection{Motivation I: Speculative Acceptance Is Domain-Dependent}

We first observe that speculative acceptance varies substantially across domains, model scales, and speculative decoding methods. Figure~\ref{fig:domain_accept_length} reports the mean number of accepted tokens per verification step for EAGLE and SAMD on Vicuna-7B and Llama-2-13B. On Vicuna-7B, SAMD consistently outperforms EAGLE, but its accept length still varies from 4.30 on translation to 5.85 on MMLU. EAGLE ranges similarly from 2.70 in translation to 3.93 in summarization. On Llama-2-13B, the relative ordering reverses: EAGLE achieves higher acceptance across all domains, reaching 5.66 on MMLU but only 3.46 on QA, while SAMD ranges from 2.53 to 3.96. These results show that acceptance is determined not only by the speculative method, but also by the interaction among the draft model, target model, and domain-specific generation task.

This variation motivates domain-specific optimization of target verification. A fixed dense verifier applies the same full-depth computation to all requests, even though different domains may require different levels of refinement. In some domains, later target transformations may unnecessarily modify task-sufficient predictions, reducing alignment with the draft model and causing earlier rejection. We therefore ask whether selectively shortening the target verification path can mitigate such domain-conditioned overthinking. This leads to our next observation: skipping selected target layers can sometimes improve both speculative acceptance and downstream task performance.
\begin{figure}[t]
  \centering
  \includegraphics[width=0.5\textwidth]{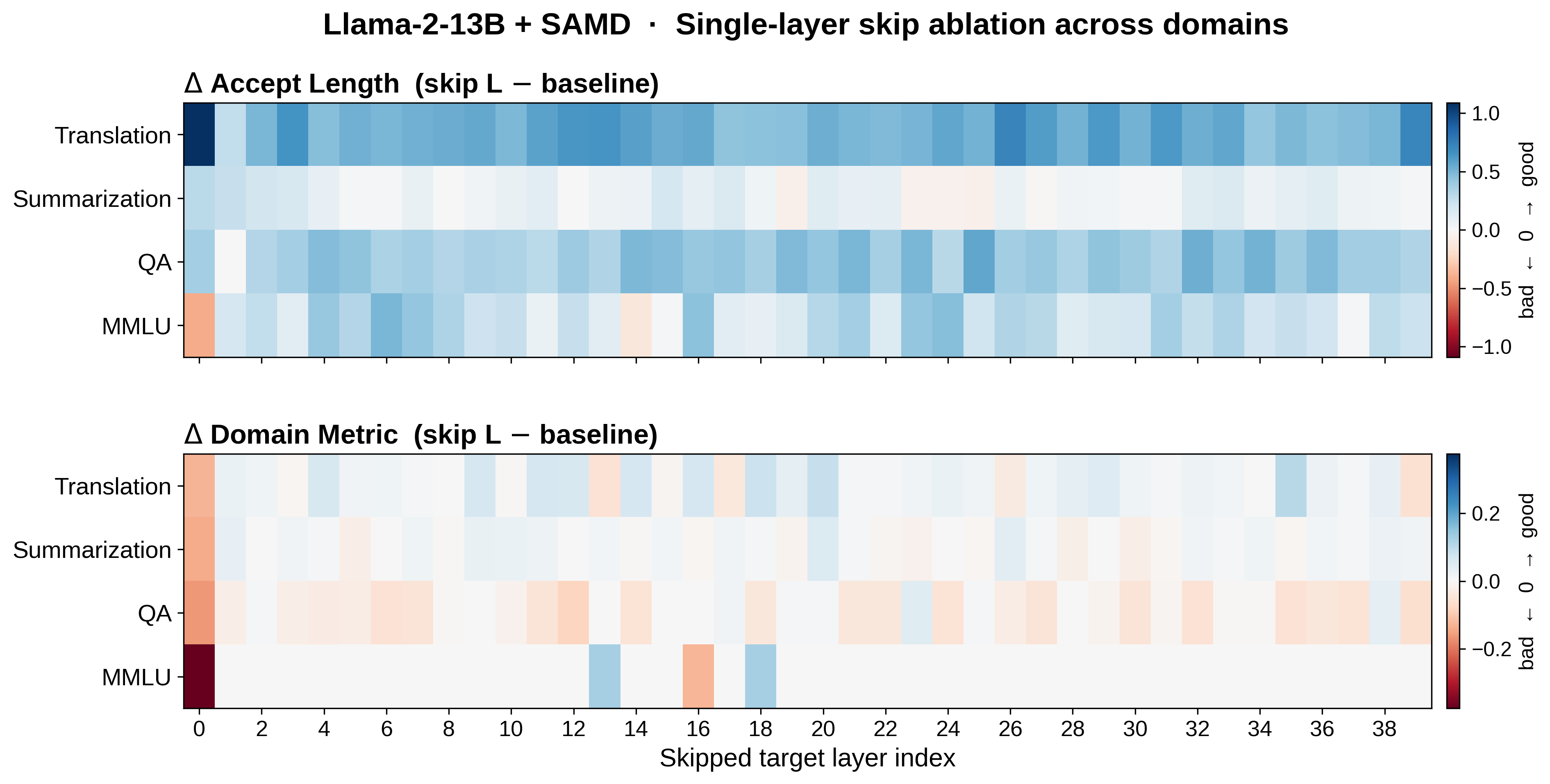}
  \vspace{-10pt}
\caption{
Single-layer skip ablation on Llama-2-13B with SAMD across four domains. 
}
\label{fig:single_layer_skip}
  \vspace{-10pt} 
\end{figure}

\begin{figure*}[!t] 
\centering 
\includegraphics[width=0.9\textwidth, trim=0 4cm 8cm 0, clip]{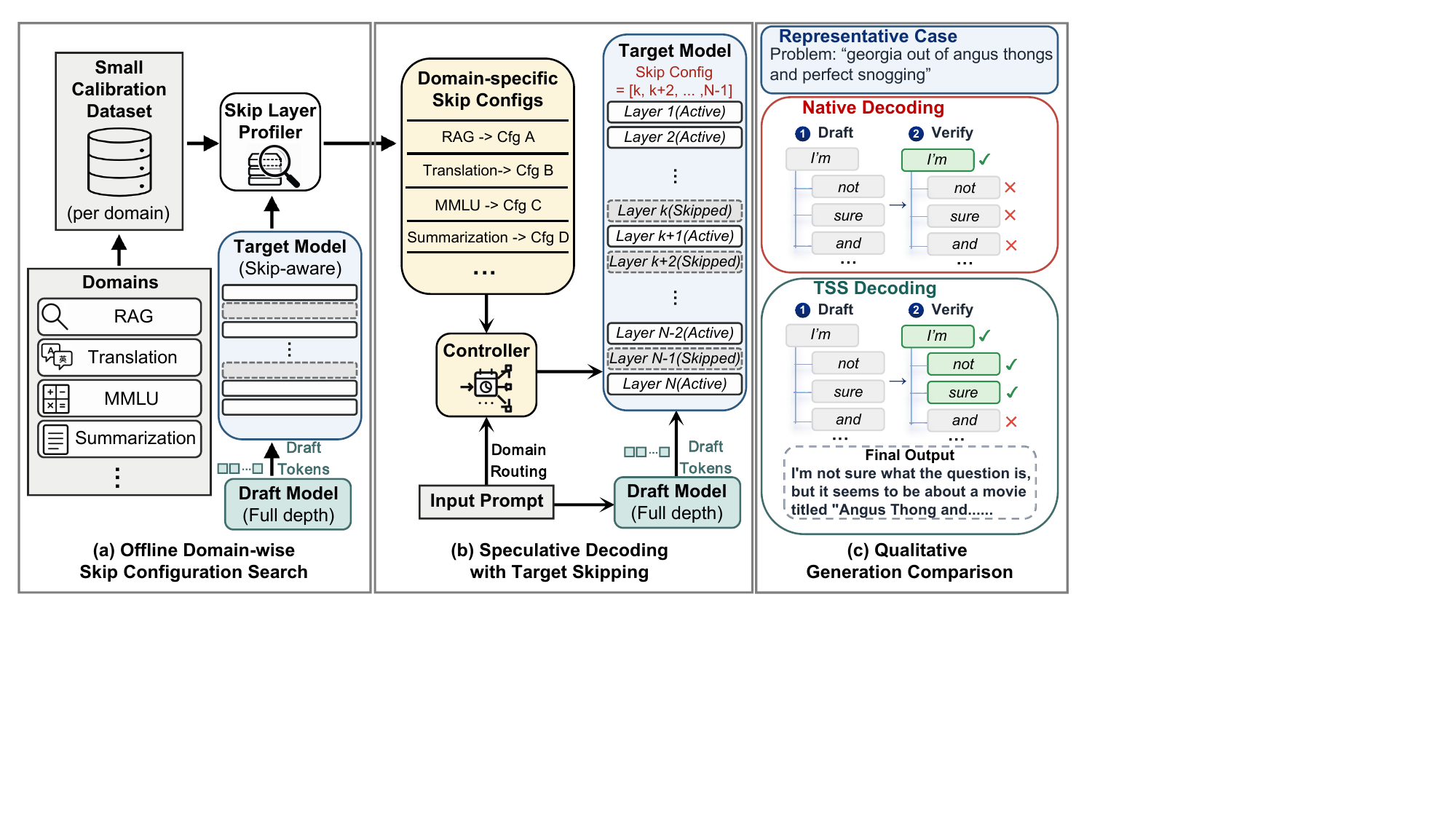} 
\vspace{-20pt} 
\caption{
Overview of TSS. (a) Offline domain-wise profiling searches target-layer skip configurations using small calibration sets. (b) At runtime, a domain-aware controller retrieves the matched configuration and executes the corresponding sparse target-verification path, while leaving the draft model unchanged. (c) A representative case shows that TSS accepts more draft tokens and improves the final output over native dense verification.
}
\label{fig:overview}
\end{figure*}
\subsection{Motivation II: Target Layers Exhibit Domain-Specific Overthinking}

To examine whether full-depth target verification may unnecessarily revise task-sufficient representations, we perform a single-layer skip ablation on Llama-2-13B with SAMD. For each target layer $l$, we skip only that layer during verification and report the change in speculative accept length and downstream task metric relative to the dense baseline. As shown in Figure~\ref{fig:single_layer_skip}, skipping many individual layers increases accept length, particularly for translation and QA. However, the magnitude of this improvement varies substantially across layers and domains. Some configurations provide little benefit, while skipping sensitive layers, such as the first target layer, can noticeably reduce acceptance in certain domains.

The downstream metrics exhibit a different and more selective pattern. Most single-layer skips produce only small metric changes, whereas a few layers improve task performance and others cause clear degradation. Moreover, the beneficial positions differ across translation, summarization, QA, and MMLU. These results suggest that full-depth verification can induce domain-conditioned overthinking: additional transformations may reduce draft--target alignment or revise predictions that are already sufficient for a particular task. At the same time, the contrasting acceptance and metric patterns show that a layer cannot be selected according to either objective alone. Effective target sparsification must jointly consider speculative acceptance and downstream task quality under each domain.

\section{Method}
\subsection{Problem Formulation}

We consider speculative decoding with a draft model $D$ and a target model $T$ serving requests from multiple domains. Given an input prefix, $D$ proposes $\gamma$ candidate tokens, which are verified in parallel by $T$. Let the target model contain $L$ Transformer blocks. For domain $d$, a target-layer skip configuration is denoted by
\[
S_d \subseteq \{1,\ldots,L\},
\]
where layers in $S_d$ are bypassed during verification, producing the sparse target verifier $T_{S_d}$. The dense verifier corresponds to $S_d=\emptyset$.

Using a small domain-specific calibration set $\mathcal{C}_d$, we evaluate each configuration according to three quantities:
\[
\begin{aligned}
A_d(S_d) &=
\operatorname{AvgAccept}(D,T_{S_d};\mathcal{C}_d),\\
M_d(S_d) &=
\operatorname{Metric}(D,T_{S_d};\mathcal{C}_d),\\
C_d(S_d) &=
\operatorname{Cost}(T_{S_d};\mathcal{C}_d),
\end{aligned}
\]
where $A_d$ is the average number of accepted draft tokens per verification step, $M_d$ is the downstream task metric, and $C_d$ is the target verification cost. Their dense-baseline values are denoted by
$A_d^0=A_d(\emptyset)$ and $M_d^0=M_d(\emptyset)$.

TSS seeks the lowest-cost domain-specific verification path that preserves both speculative acceptance and task performance:
\[
\begin{aligned}
S_d^\star
=
\operatorname*{arg\,min}_{S_d\subseteq\{1,\ldots,L\}}
\quad & C_d(S_d)\\
\text{s.t.}\quad
& A_d(S_d)\geq A_d^0-\epsilon_A,\\
& M_d(S_d)\geq M_d^0-\epsilon_M,
\end{aligned}
\]
where $\epsilon_A$ and $\epsilon_M$ tolerate small calibration fluctuations. This formulation jointly constrains draft-target alignment and domain-level quality, preventing acceleration from being achieved at the expense of either objective.

As illustrated in Figure~\ref{fig:overview}, TSS consists of an offline configuration stage and an online execution stage. Offline, a skip-layer profiler evaluates the target model on small domain-specific calibration sets and stores the selected skip masks in a domain-to-configuration mapping. Online, a lightweight controller routes each request to its matched configuration, while the draft model remains unchanged and the target verifier bypasses the selected layers. The qualitative example further shows that the resulting sparse verifier can accept more draft tokens and produce a better task-level output than native dense verification.
\begin{algorithm}[t]
\caption{Acceptance- and Metric-Aware Breadth Search}
\label{alg:tss_breadth_search}
\begin{algorithmic}[1]
\Require $D,T,\mathcal{C}_d,L,K$ and thresholds
$\epsilon_A,\epsilon_M,\tau_A,\tau_M$
\Ensure Best configuration $S_d^\star$ and Pareto set $\mathcal{P}_d$

\State Evaluate dense baseline $A_d^0,M_d^0,C_d^0$
\State $\mathcal{L}_d\gets\emptyset$

\For{$l=1$ to $L$}
    \State Evaluate $S=\{l\}$
    \If{$A_d(S)\ge A_d^0-\tau_A
        \lor M_d(S)\ge M_d^0-\tau_M$}
        \State $\mathcal{L}_d\gets\mathcal{L}_d\cup\{l\}$
    \EndIf
\EndFor

\State $\mathcal{F}_d\gets\{\emptyset\}$,
$\mathcal{P}_d\gets\{(\emptyset,A_d^0,M_d^0,C_d^0)\}$

\For{$k=1$ to $K$}
    \State $\mathcal{S}_k\gets
    \{S\subseteq\mathcal{L}_d:|S|=k\}$
    \State $\mathcal{F}_d^{(k)}\gets\emptyset$

    \ForAll{$S\in\mathcal{S}_k$}
        \State Evaluate $A_d(S),M_d(S),C_d(S)$
        \If{$A_d(S)\ge A_d^0-\epsilon_A
            \land M_d(S)\ge M_d^0-\epsilon_M$}
            \State $\mathcal{F}_d^{(k)}
            \gets\mathcal{F}_d^{(k)}\cup\{S\}$
            \State Add $(S,A_d(S),M_d(S),C_d(S))$ to $\mathcal{P}_d$
        \EndIf
    \EndFor

    \If{$\mathcal{F}_d^{(k)}=\emptyset$}
        \State \textbf{break}
    \EndIf
    \State $\mathcal{F}_d\gets\mathcal{F}_d\cup\mathcal{F}_d^{(k)}$
\EndFor

\State Remove dominated configurations from $\mathcal{P}_d$
\State $S_d^\star\gets
\arg\min_{S\in\mathcal{F}_d}C_d(S)$
\State \Return $S_d^\star,\mathcal{P}_d$
\end{algorithmic}
\end{algorithm}

\subsection{Acceptance- and Metric-Aware Breadth-First Search}

Selecting multiple target layers is challenging because speculative acceptance and downstream task performance do not always change in the same direction. A layer that improves the task metric may slightly reduce acceptance, while another layer may improve acceptance but provide little task-level benefit. Consequently, assigning a fixed priority or combining the two objectives into a weighted score can prematurely bias the search toward one signal. Conventional greedy search is particularly vulnerable to this issue: once it commits to a locally favorable layer, all subsequent configurations are restricted to descendants of that choice, potentially missing combinations formed by individually moderate but jointly beneficial layers. Moreover, the effect of skipping multiple Transformer blocks is not necessarily additive, since removing one block can alter the representations received by all subsequent blocks.

Algorithm~\ref{alg:tss_breadth_search} presents our acceptance- and metric-aware breadth-first search. The algorithm first evaluates the native dense verifier to establish the baseline acceptance, downstream metric, and verification cost. It then independently evaluates every single-layer skip and constructs a broad candidate pool:
\[
\mathcal{L}_d
=
\left\{
l \;\middle|\;
A_d(\{l\}) \geq A_d^0-\tau_A
\;\lor\;
M_d(\{l\}) \geq M_d^0-\tau_M
\right\}.
\]
The logical \emph{or} retains a layer when it is promising under either acceptance or the downstream metric. This permissive initial filtering is important because a layer that appears unfavorable under one objective in isolation may still become useful when combined with other layers. Unlike greedy ranking, this stage does not collapse acceptance and task performance into a single scalar score or discard candidates according to a fixed objective priority.

After constructing $\mathcal{L}_d$, TSS explores multi-layer configurations level by level. At search depth $k$, the algorithm enumerates all unique configurations that skip exactly $k$ candidate layers:
\[
\mathcal{S}_k
=
\left\{
S\subseteq\mathcal{L}_d
\;\middle|\;
|S|=k
\right\}.
\]
Thus, all configurations with the same sparsity level are evaluated before proceeding to configurations with more skipped layers. This breadth-first organization allows combinations originating from different single-layer choices to compete at the same search depth, avoiding the early commitment of greedy search. For every configuration, TSS measures acceptance, downstream performance, and target verification cost on the domain-specific calibration set.

A configuration is retained only when it jointly satisfies both feasibility requirements:
\[
\mathcal{F}_d^{(k)}
=
\left\{
S\in\mathcal{S}_k
\;\middle|\;
\begin{aligned}
A_d(S) &\geq A_d^0-\epsilon_A,\\
M_d(S) &\geq M_d^0-\epsilon_M
\end{aligned}
\right\}.
\]
The initial candidate construction therefore uses a permissive \emph{or} condition to preserve search diversity, whereas the multi-layer evaluation uses a strict \emph{and} condition to prevent acceleration from sacrificing either speculative acceptance or task quality. The search continues until the maximum depth $K$ is reached or the current depth contains no feasible configuration.

Finally, TSS removes configurations dominated in both acceptance and downstream metric and selects the feasible configuration with the lowest measured verification cost:
\[
S_d^\star
=
\arg\min_{S\in\bigcup_{k=0}^{K}\mathcal{F}_d^{(k)}}
C_d(S).
\]
The selected configuration becomes the default skip policy for domain $d$. Other Pareto-optimal configurations are also retained, allowing the domain-to-configuration mapping to support alternative operating points when different efficiency or quality requirements are preferred. Since the search is performed offline, its additional evaluation cost does not affect the runtime speculative decoding process.
\subsection{Domain-to-Configuration Mapping}

The optimal target-layer skip pattern varies across domains, because different tasks exhibit different acceptance behavior and layer redundancy. Therefore, TSS does not use a single global skip mask. Instead, after offline breadth search, it constructs a domain-specific configuration map:
\[
\Phi:\mathcal{D}\rightarrow 2^{\{1,\ldots,L\}},
\qquad
\Phi(d)=S_d^\star,
\]
where $\mathcal{D}$ is the set of supported domains and $S_d^\star$ is the selected skip-layer set for domain $d$.

Each configuration stores the information required for runtime execution:
\[
\mathcal{G}_d
=
\left(
S_d^\star,\,
A_d(S_d^\star),\,
M_d(S_d^\star),\,
R_d
\right),
\qquad
R_d=\frac{|S_d^\star|}{L},
\]
where $A_d(S_d^\star)$ and $M_d(S_d^\star)$ are the calibrated accept length and task metric, and $R_d$ is the target-layer sparsity ratio. The resulting mapping is saved as a lightweight local configuration file. At runtime, the controller retrieves $\mathcal{G}_d$ according to the request domain and applies the corresponding skip set to the target verifier. This allows one complete target model to support multiple domain-specific sparse paths without storing separate pruned model copies.

\begin{table*}[t]
\centering
\small
\caption{The results on Spec-Bench.}
\label{tab:main_tss_results}
\renewcommand{\arraystretch}{0.3}
\setlength{\tabcolsep}{5pt}
\begin{tabular}{l|l|cccccc}
\toprule
Domain & Method & Accept Len. & Task Metric & Skip Layers &
Sparsity & Throughput(tok/s) & vs.\ Native \\
\midrule
\multicolumn{8}{l}{\textbf{Vicuna-7B}
(EAGLE / EAGLE+TSS, $L=32$)} \\
\midrule
Translation
& EAGLE
& 2.70 & 0.131 & -- & -- & 75.6 & 1.00$\times$ \\
& EAGLE+TSS
& \textbf{4.53} & \textbf{0.237}
& $\{3,25,30\}$ & 9.4\%
& \textbf{127.3} & \textbf{1.68$\times$} \\
\midrule
Summarization
& EAGLE
& 3.93 & 0.266 & -- & -- & 89.4 & 1.00$\times$ \\
& EAGLE+TSS
& \textbf{4.01} & \textbf{0.270}
& $\{15,22,23\}$ & 9.4\%
& \textbf{95.5} & \textbf{1.07$\times$} \\
\midrule
RAG
& EAGLE
& 3.24 & 0.096 & -- & -- & 76.0 & 1.00$\times$ \\
& EAGLE+TSS
& \textbf{4.03} & \textbf{0.109}
& $\{3,6,14,25,30\}$ & 15.6\%
& \textbf{97.1} & \textbf{1.28$\times$} \\
\midrule
QA
& EAGLE
& 2.93 & 0.042 & -- & -- & 80.0 & 1.00$\times$ \\
& EAGLE+TSS
& \textbf{3.97} & \textbf{0.064}
& $\{6,8,19,21,27\}$ & 15.6\%
& \textbf{113.7} & \textbf{1.42$\times$} \\
\midrule
MMLU
& EAGLE
& 3.12 & 0.281 & -- & -- & 67.5 & 1.00$\times$ \\
& EAGLE+TSS
& \textbf{3.96} & \textbf{0.328}
& $\{3,7,9,14,20\}$ & 15.6\%
& \textbf{90.8} & \textbf{1.35$\times$} \\

\midrule
\multicolumn{8}{l}{\textbf{Llama-2-13B}
(SAMD Token-Recycle / SAMD+TSS, $L=40$)} \\
\midrule
Translation
& SAMD
& 2.69 & 0.208 & -- & -- & 38.3 & 1.00$\times$ \\
& SAMD+TSS
& \textbf{2.89} & \textbf{0.212}
& $\{7,11,20,31,35,38\}$ & 15.0\%
& \textbf{49.6} & \textbf{1.29$\times$} \\
\midrule
Summarization
& SAMD
& 3.11 & 0.249 & -- & -- & 45.8 & 1.00$\times$ \\
& SAMD+TSS
& \textbf{3.19} & \textbf{0.251}
& $\{12,20,27,33,34,38\}$ & 15.0\%
& \textbf{53.2} & \textbf{1.16$\times$} \\
\midrule
QA
& SAMD
& 2.53 & 0.120 & -- & -- & 33.2 & 1.00$\times$ \\
& SAMD+TSS
& \textbf{2.74} & \textbf{0.129}
& $\{9,10,11,19,28,38\}$ & 15.0\%
& \textbf{43.3} & \textbf{1.30$\times$} \\
\midrule
MMLU
& SAMD
& 3.96 & 0.359 & -- & -- & 58.6 & 1.00$\times$ \\
& SAMD+TSS
& \textbf{4.08} & \textbf{0.422}
& $\{7,23,24,26,28,32,33\}$ & 17.5\%
& \textbf{70.8} & \textbf{1.21$\times$} \\
\bottomrule
\end{tabular}
\end{table*}

\subsection{Domain-Aware Skip Controller}

At runtime, TSS retains the complete target model and applies domain-specific sparsity through a lightweight skip controller rather than permanent parameter pruning. Given an input request $x$, the controller first identifies its domain $d(x)$ and queries the offline domain-to-configuration mapping:
\[
S(x)=
\begin{cases}
\Phi(d(x)), & d(x)\in\mathcal{D},\\
\emptyset, & d(x)\notin\mathcal{D},
\end{cases}
\]
where $\mathcal{D}$ denotes the set of supported domains. For a supported domain, the retrieved skip configuration remains fixed throughout the request. For an unseen or unsupported domain, the controller falls back to $S(x)=\emptyset$, and the target model executes its original full-depth verification path. The draft model and candidate proposal process remain unchanged in both cases.

Before executing each Transformer block, the target forward function checks whether its layer index belongs to the selected skip set:
\[
h_{l+1}=
\begin{cases}
f_l(h_l), & l\notin S(x),\\
h_l, & l\in S(x).
\end{cases}
\]
A skipped block therefore forwards its input hidden state directly to the next active block, avoiding its attention and feed-forward computation. Since all target parameters remain loaded, requests can switch among domain-specific sparse paths or the dense fallback path without loading separate pruned models. The controller only performs a simple domain lookup followed by a match check on each layer index, making its runtime overhead practically negligible compared with the computation saved by bypassing Transformer blocks. Thus, TSS provides efficient domain-aware execution while preserving the original dense target behavior for requests outside the configured domain set.

\section{Experiments}

\subsection{Experimental Setup}

We evaluate TSS on the domain-specific tasks provided by Spec-Bench. For each domain, we split the corresponding data into a calibration set and a test set, using 20\% of the prompts for target-layer skip selection and the remaining 80\% for held-out evaluation. The calibration set is used only to determine the domain-specific sparse target configuration, and no model weights are updated. In all experiments, we set both $\epsilon$ and $\tau$ to $5\%$. Experimental costs and hardware configurations are detailed in technical supplement.

\paragraph{Models and speculative decoding methods.}
We evaluate TSS on two representative speculative decoding settings: Vicuna-7B with EAGLE~\cite{li2024eagle} and Llama-2-13B with SAMD Token-Recycle~\cite{SAMD}. Vicuna-7B contains $L=32$ target layers, while Llama-2-13B contains $L=40$ target layers. For each setting, \textbf{Native} denotes the original speculative decoding method with the dense target verifier, and \textbf{+TSS} denotes the same method equipped with target-layer skipping. TSS keeps the draft model, candidate generation procedure, and accept/reject logic unchanged; only the target verifier is replaced by a domain-calibrated sparse verifier.

\paragraph{Domains and metrics.}
We evaluate on four tasks from Spec-Bench~\cite{xia-etal-2024-unlocking}—translation, summarization, QA, and RAG—and additionally include MMLU as a multiple-choice knowledge benchmark. For translation, we report BLEU, which measures $n$-gram precision between the generated translation and reference translations with a brevity penalty~\cite{papineni2002bleu}. For summarization, we report ROUGE-L F1, which evaluates the longest common subsequence overlap between the generated summary and the reference summary~\cite{lin2004rouge}. For RAG and QA, we follow the standard open-domain question answering evaluation and report token-level F1 after answer normalization, measuring the overlap between the predicted answer and the reference answer~\cite{kwiatkowski2019naturalquestions,rajpurkar2016squad}. For MMLU, we report multiple-choice accuracy over the predicted answer option, following the original MMLU evaluation protocol~\cite{hendrycks2021mmlu}. 
\subsection{Main Results}

Table~\ref{tab:main_tss_results} reports the main results on Spec-Bench. Across all evaluated domains and model settings, TSS consistently improves throughput over the native speculative decoding baseline while maintaining or improving the downstream task metric. On Vicuna-7B with EAGLE, TSS achieves speedups from 1.07$\times$ to 1.68$\times$ across five domains. The largest gain appears on translation, where TSS increases the accept length from 2.70 to 4.53 and improves throughput from 75.6 to 127.3 tok/s, corresponding to a 1.68$\times$ speedup. Similar improvements are observed on RAG, QA, and MMLU, where TSS improves both accept length and task metric while skipping 15.6\% of target layers.

TSS also generalizes to larger target models and different speculative decoding methods. On Llama-2-13B with SAMD Token-Recycle, TSS improves throughput by 1.16$\times$--1.30$\times$ across translation, summarization, QA, and MMLU. Importantly, these speedups do not come at the cost of task performance. For example, on MMLU, TSS improves the task metric from 0.359 to 0.422 while increasing throughput from 58.6 to 70.8 tok/s. On QA, TSS improves accept length from 2.53 to 2.74 and task metric from 0.120 to 0.129, with a 1.30$\times$ throughput gain. These results show that target-side sparsification is not limited to a specific draft mechanism or model size.

The magnitude of TSS gains varies across speculative decoding architectures. Vicuna-7B with EAGLE exhibits larger improvements, while Llama-2-13B with SAMD Token-Recycle shows more moderate but consistent gains. This difference suggests that the benefit of target-side skipping depends on the interaction among the draft mechanism, target model, and domain, rather than model scale alone. Nevertheless, TSS improves throughput across both architectures while preserving or improving downstream metrics, demonstrating that target-side optimization is broadly applicable to heterogeneous speculative decoding pipelines.

\begin{table}[t]
\centering
\small
\caption{
Comparison of target-layer search strategies on MMLU and Translation, SLEB~\cite{song2024sleb} and GM-Skip~\cite{huang2026gmskip}.
}
\label{tab:search_ablation}
\renewcommand{\arraystretch}{0.92}
\setlength{\tabcolsep}{1pt}
\begin{tabular}{l|cccc}
\toprule
Method & Accept Len. & Metric & Skip Layers & Throughput \\
\midrule
\multicolumn{5}{l}{\textit{Translation}} \\
\midrule
SLEB
& 3.477 & 0.129 & $\{2,3,4,9\}$ & 101.5 \\
Random
& 3.382 & 0.140 & $\{2,3,4,5\}$ & 100.3 \\
GM-Skip
& 2.995 & 0.109 & $\{12,14,22,23\}$ & 91.1 \\
Accept-only
& 3.769 & 0.126 & $\{3,15\}$ & 107.9 \\
Metric-only
& 3.029 & 0.139 & $\{12,18,28\}$ & 91.1 \\
TSS
& \textbf{4.528} & \textbf{0.237}
& $\{3,25,30\}$ & \textbf{127.3} \\
\midrule
\multicolumn{5}{l}{\textit{MMLU}} \\
\midrule
SLEB
& 3.488 & 0.296 & $\{2,3,4,6\}$ & 84.0 \\
Random
& 3.539 & 0.132 & $\{2,11,17\}$ & 84.9 \\
GM-Skip
& 2.889 & 0.311 & $\{3,8,19\}$ & 70.4 \\
Accept-only
& \textbf{4.023} & 0.219 & $\{4,6,16\}$ & 88.0 \\
Metric-only
& 3.001 & 0.265 & $\{3\}$ & 74.1 \\
TSS
& 3.955 & \textbf{0.328}
& $\{3,7,9,14,20\}$ & \textbf{90.8} \\
\bottomrule
\end{tabular}
\end{table}

Overall, TSS achieves simultaneous gains in speculative acceptance, task quality, and end-to-end generation speed. Averaged over all settings in Table~\ref{tab:main_tss_results}, TSS improves throughput by approximately $1.31\times$, and increases accept length by about 20\% over the native baseline. These results support our central claim that full-depth verification can induce domain-conditioned overthinking, and that selectively bypassing unnecessary target transformations can improve both draft--target alignment and inference efficiency. All results can be reproduced on Spec-Bench by directly applying the reported skip-layer configurations without additional training or model modification.
\begin{figure}[t]
  \centering
  \includegraphics[width=0.5\textwidth,trim=0cm 9cm 10cm 0cm, clip]{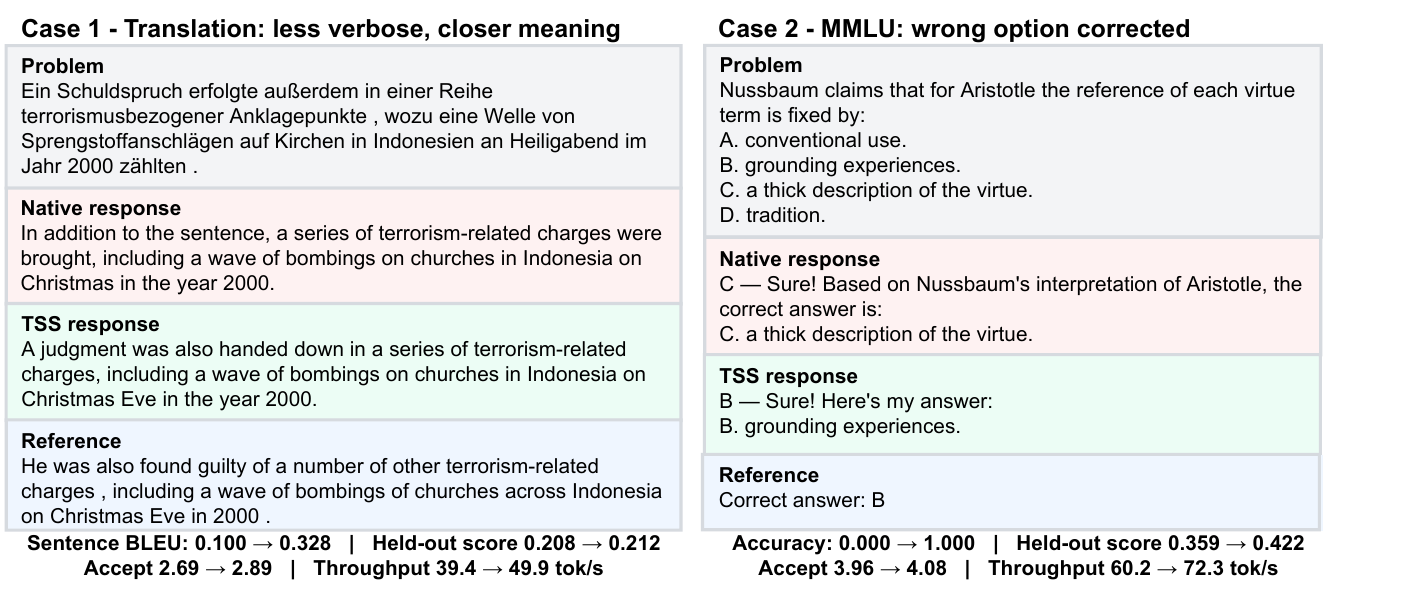}
  \vspace{-10pt}
\caption{
Representative translation and MMLU cases. TSS produces an output closer to the translation reference and corrects an incorrect MMLU answer, while simultaneously improving accept length and throughput.
}
\label{fig:case_study}
  \vspace{-10pt} 
\end{figure}

\subsection{Search Algorithm Ablation}
Table~\ref{tab:search_ablation} compares TSS with several alternative
target-layer selection strategies. On Translation, TSS Breadth Search
achieves the best results across all three objectives, increasing the
accept length to 4.528, the task metric to 0.237, and throughput to
127.3 tokens/s. In contrast, GM-Skip(Greedy search) and random search select
configurations that degrade both acceptance and task quality, while
the single-objective variants improve only one aspect.

On MMLU, accept-only search obtains the highest accept length, but
substantially reduces accuracy from 0.328 to 0.219. Accept-only search
matches the best task accept, yet produces much lower metrics and
throughput. TSS reaches the same best accuracy while maintaining an
accept length of 3.955 and achieving the highest throughput of
90.8 tokens/s. These results show that optimizing acceptance or task
quality independently is insufficient, and that breadth search provides
a better balance among acceptance, downstream performance, and
inference efficiency.

\subsection{Case Study}

Figure~\ref{fig:case_study} presents two representative examples where TSS improves both generation quality and speculative decoding efficiency. In the translation example, the native model produces a verbose sentence with several expressions that deviate from the reference. After applying TSS, the output becomes more concise and semantically closer to the reference translation. Consequently, sentence-level BLEU increases from 0.100 to 0.328, while the average accept length improves from 2.69 to 2.89 and throughput increases from 39.4 to 49.9 tokens/s.

The MMLU example shows a stronger effect on task correctness. The native speculative model selects option C, whereas TSS produces option B, which matches the reference answer. This changes the sample-level accuracy from 0 to 1 and is consistent with the overall MMLU score improvement from 0.359 to 0.422. At the same time, TSS increases the accept length from 3.96 to 4.08 and throughput from 60.2 to 72.3 tokens/s. These examples suggest that target-layer skipping does not simply weaken verification. By removing domain-redundant computation, TSS can avoid unnecessary output refinement, improve draft-target alignment, and produce outputs that better match downstream task objectives.
\section{Conclusion}

We presented \textbf{TSS}, a target-side sparsification framework for domain-specific speculative decoding. TSS uses acceptance- and metric-aware search to identify domain-specific skip configurations and applies them through a lightweight controller without retraining or separate pruned models. Experiments on Spec-Bench show consistent gains in accept length and throughput while preserving or improving task metrics across multiple domains and speculative decoding methods. These results demonstrate that full-depth target verification is not always optimal and establish target-side optimization as a complementary direction for speculative decoding. Additional ablations and qualitative analyses for all five domains are provided in the technical supplement.

\bibliography{aaai2027}

\clearpage
\appendix

\section{Experimental Setup}
\label{sec:setup}

\subsection{Models}
\label{sec:setup-models}

We evaluate target-layer skipping (TSS) on two speculative-decoding stacks that cover both mid-size and larger chat models.

\paragraph{Vicuna-7B + EAGLE.}
The target model is \texttt{lmsys/vicuna-7b-v1.3}, a LLaMA-based chat model with $L{=}32$ Transformer layers, hidden size $4096$, $32$ attention heads, intermediate size $11008$, vocabulary size $32{,}000$, and context length $2048$.
Weights are loaded in \texttt{float16}.
The draft model is the official EAGLE checkpoint \texttt{EAGLE-Vicuna-7B-v1.3} (a one-layer autoregressive draft head aligned to the Vicuna-7B representation), used in the original EAGLE (non-EAGLE-3) configuration.
Prompts follow the Vicuna conversation template.

\paragraph{Llama-2-13B + SAMD.}
The target model is \texttt{meta-llama/Llama-2-13b-chat-hf} with $L{=}40$ layers, hidden size $5120$, $40$ attention heads, intermediate size $13824$, vocabulary size $32{,}000$, and context length $4096$ (\texttt{float16}).
Speculative decoding uses SAMD with \emph{Token Recycle} drafting and tree configuration \texttt{token\_recycle\_4\_15} (Spec-Bench setting).
Prompts use the Llama-2 chat template (\texttt{[INST]}~\ldots~\texttt{[/INST]}).

\subsection{Hardware and Software}
\label{sec:setup-hardware}

All measurements are obtained on 8*NVIDIA GeForce RTX~4090 GPU with $24$\,GB GDDR6X memory (driver $595.71$, compute capability $8.9$).
The host has dual-socket Intel Xeon Platinum~8470Q CPUs ($2{\times}52$ cores / $208$ threads) and $754$\,GB system RAM, running Ubuntu~$22.04$ LTS.
We use PyTorch~$2.11$ with CUDA~$12.8$/$13.0$ and Hugging Face Transformers ($4.46$ for the EAGLE stack; $4.57$ for the SAMD stack). models are loaded in \texttt{float16} on device.

\subsection{Evaluation Protocol}
\label{sec:setup-eval}

\begin{itemize}
  \item \textbf{Datasets.} Spec-Bench-style domains: translation (BLEU), summarization (ROUGE-L), open-domain QA (token F1), RAG (token F1), and MMLU (accuracy).
  (\emph{test}). Exception: Llama-2-13B QA selection uses $32$ train prompts.
  \item \textbf{Generation.} Maximum new tokens per request $T{=}96$; decoding temperature $\tau{=}0$ (greedy); EAGLE draft budget \texttt{total\_token}${=}60$.
  \item \textbf{Metrics.} Mean accepted length per verify step, task metric, end-to-end throughput (tok/s), skip set $\mathcal{S}$ and sparsity $|\mathcal{S}|/L$.
  \item \textbf{Native vs.\ TSS.} \emph{Native}: $\mathcal{S}{=}\emptyset$. \emph{+TSS}: selected target-layer skip set during verification.
\end{itemize}

\begin{table*}[t]
\centering
\small
\caption{Default hyperparameters for TSS selection and decoding.}
\label{tab:hparams}
\setlength{\tabcolsep}{4pt}
\begin{tabular}{llc}
\toprule
Category & Setting & Value \\
\midrule
\multirow{4}{*}{Decoding}
  & Precision & \texttt{float16} \\
  & Temperature $\tau$ & $0$ \\
  & Max new tokens $T$ & $96$ \\
  & EAGLE \texttt{total\_token} & $60$ \\
\midrule
\multirow{4}{*}{Data}
  & Random seed & $42$ \\
  & Selection train size & $16$ (QA-13B: $32$) \\
  & Held-out test size & $64$ \\
  & Prompt pool & $80$ / domain \\
\midrule
\multirow{8}{*}{TSS search}
  & Beam width & $3$ \\
  & Early barrier & $2$ \\
  & Latter barrier & $0$ \\
  & Accept constraint (explore) & accept $\ge$ native (hard) \\
  & Score tolerance (relative) & drop $\le 5\%$ \\
  & Min skips (final prefer) & $\min(3,|\mathcal{S}|_{\max})$ \\
\midrule
\multirow{3}{*}{Selection rule}
  & Triple-win (main) & score, accept, tok/s $\ge$ native \\
  & Final ranking & accept $\uparrow$, then $|\mathcal{S}|$ $\uparrow$, then tok/s $\uparrow$ \\
  & Layers preferred & latter $\rightarrow$ early \\
\bottomrule
\end{tabular}
\end{table*}

\begin{table*}[t]
\caption{
Sensitivity of TSS to the acceptance and metric tolerances on Vicuna-7B across four domains. We jointly vary $\epsilon_A$ and $\epsilon_M$ and report the resulting accept length, downstream metric, skip configuration, sparsity, throughput, and search evaluations. The $-5\%$ setting provides the best overall balance between task quality and speculative acceptance.
}
\label{tab:tolerance_sensitivity}
\renewcommand{\arraystretch}{0.90}
\setlength{\tabcolsep}{3.5pt}
\begin{tabular*}{\textwidth}{
@{\extracolsep{\fill}}
llccccc c
@{}
}
\toprule
Domain & Score Tol. & Accept Len. & Metric & Skip Layers & Sparsity & Tok/s & Evals \\
\midrule

\multirow{5}{*}{Translation}
& $+5\%$  & 4.134 & 0.1207 & $\{4,9\}$       & 12.5\% & 120.7 & 351 \\
& $0\%$   & 4.308 & 0.1402 & $\{3,11,14,15\}$      & 12.5\% & 122.8 & 349 \\
& \textbf{$-5\%$}
& \textbf{4.528} & \textbf{0.2370}
& \textbf{\{3,25,30\}} & \textbf{9.4\%}
& \textbf{127.3} & -- \\
& $-10\%$ & 4.308 & 0.1402 & $\{3,11,14,15\}$      & 12.5\% & 122.8 & 348 \\
& $-20\%$ & 4.289 & 0.1462 & $\{3,15,23,24\}$      & 12.5\% & 125.5 & 347 \\
\midrule

\multirow{5}{*}{QA}
& $+5\%$  & 3.726 & 0.0600 & $\{19,27\}$       & 12.5\% & 100.1 & 354 \\
& $0\%$   & 3.801 & 0.0600 & $\{3,14,22,25,30\}$   & 15.6\% & 107.3 & 349 \\
& \textbf{$-5\%$}
& \textbf{3.970} & \textbf{0.0640}
& \textbf{\{6,8,19,21,27\}} & \textbf{15.6\%}
& \textbf{113.7} & -- \\
& $-10\%$ & 3.801 & 0.0600 & $\{3,14,22,25,30\}$   & 15.6\% & 107.3 & 350 \\
& $-20\%$ & 3.781 & 0.0415 & $\{3,11,30\}$         & 9.4\%  & 104.8 & 352 \\
\midrule

\multirow{5}{*}{RAG}
& $+5\%$  & 4.374 & 0.0432 & $\{2,11,29\}$     & 15.6\% & 102.4 & 351 \\
& $0\%$   & 4.374 & 0.0432 & $\{2,3,4,11,29\}$     & 15.6\% & 102.4 & 352 \\
& \textbf{$-5\%$}
& \textbf{4.027} & \textbf{0.1091}
& \textbf{\{3,6,14,25,30\}} & \textbf{15.6\%}
& \textbf{97.1} & -- \\
& $-10\%$ & 3.990 & 0.0824 & $\{4,5,11,12,29\}$    & 15.6\% & 97.7  & 351 \\
& $-20\%$ & 4.428 & 0.0826 & $\{4,7,11,12,29\}$    & 15.6\% & 106.5 & 351 \\
\midrule

\multirow{5}{*}{MMLU}
& $+5\%$  & 3.115 & 0.2812 & $\emptyset$           & 0.0\%  & 70.5 & 30 \\
& $0\%$   & 3.572 & 0.2969 & $\{5,11,29\}$    & 15.6\% & 82.5 & 352 \\
& \textbf{$-5\%$}
& \textbf{3.955} & \textbf{0.3281}
& \textbf{\{3,7,9,14,20\}} & \textbf{15.6\%}
& \textbf{90.8} & -- \\
& $-10\%$ & 3.972 & 0.2969 & $\{5,8,11,23,29\}$    & 15.6\% & 92.5 & 352 \\
& $-20\%$ & 3.972 & 0.2969 & $\{5,8,11,23,29\}$    & 15.6\% & 92.5 & 352 \\
\bottomrule
\end{tabular*}
\end{table*}
\subsection{Sensitivity to Acceptance and Metric Tolerances}

We study the sensitivity of TSS to the feasibility thresholds $\epsilon_A$ and $\epsilon_M$ by varying them jointly across five settings, as reported in Table~\ref{tab:tolerance_sensitivity}. The score tolerance denotes the permitted change relative to the dense baseline: $0\%$ requires both acceptance and the task metric to remain no lower than the baseline, while $-5\%$, $-10\%$, and $-20\%$ allow progressively larger decreases. In contrast, $+5\%$ requires both quantities to improve by at least $5\%$. Different tolerances lead to different skip configurations, illustrating that the thresholds directly control the balance between acceptance, task quality, and acceleration.

The $-5\%$ setting provides the most consistent balance across domains. On Translation, it achieves the highest accept length of 4.528 and the highest metric of 0.2370, together with the best throughput of 127.3 tokens/s. On QA, it similarly obtains the best acceptance, metric, and throughput, reaching 3.970, 0.0640, and 113.7 tokens/s, respectively. On RAG, more permissive settings can produce higher acceptance; for example, $-20\%$ reaches 4.428 accepted tokens. However, its task metric is only 0.0826, compared with 0.1091 under $-5\%$. This result demonstrates that maximizing acceptance alone may select configurations that accept more draft tokens but provide weaker downstream quality. On MMLU, $-10\%$ and $-20\%$ slightly increase accept length from 3.955 to 3.972 and throughput from 90.8 to 92.5 tokens/s, but reduce accuracy from 0.3281 to 0.2969. The small efficiency gain therefore comes at a clear cost in task performance.

Overall, the $-5\%$ tolerance achieves the highest downstream metric in all four evaluated domains and the highest mean accept length across the tested settings. Its average throughput is also nearly identical to the most permissive setting, while avoiding the metric degradation observed under larger tolerances. We therefore use $\epsilon_A=\epsilon_M=5\%$ throughout the main experiments, corresponding to the $-5\%$ row in the table. This setting allows small calibration fluctuations while maintaining stable task performance and selecting configurations with strong speculative acceptance.
\subsection{Qualitative Case Analysis}

Table~\ref{tab:tss_case_index} summarizes representative held-out cases across domains and model scales, while Table~\ref{tab:tss_cases_7b} provides detailed comparisons for Vicuna-7B with EAGLE. The examples exhibit three outcomes of domain-specific target skipping: TSS may improve the task-level output, preserve the original task decision while changing the generation behavior, or slightly degrade an individual response. This variation is expected because each skip configuration is selected according to aggregate acceptance and task performance on a domain-level calibration set, rather than optimized separately for every prompt. The cases therefore illustrate both the potential benefits and the request-level limitations of TSS.

The Translation examples demonstrate this variability clearly. In the 7B case, the native output closely matches the reference, whereas TSS omits the final word ``value.'' Although the meaning remains largely unchanged, this omission decreases the case-level BLEU score from 0.142 to 0.101. The degradation is relatively small and represents a case where several skipped layers remain useful for the precise surface realization of an individual prompt. In contrast, the 13B Translation example exhibits the opposite behavior: TSS produces an output closer to the reference and increases BLEU from 0.100 to 0.328. Together, these cases show that a domain-level skip configuration does not improve every translation instance uniformly, but can provide a favorable aggregate trade-off even when a small number of local degradations occur.

The Summarization cases show more consistent improvements. For the 7B example, both Native and TSS capture the main event involving Arsenal and Petr Cech, but the TSS output preserves more reference-relevant content and avoids part of the unnecessary continuation produced by Native. Consequently, ROUGE-L increases from 0.281 to 0.430. The corresponding 13B example similarly improves from 0.143 to 0.280. These cases suggest that selected target layers may introduce additional wording or refinements that reduce overlap with concise reference summaries. Bypassing these layers can preserve a generation path that better matches the summarization objective.

A similar pattern appears in open-domain QA. In the 7B example, Native produces a broad explanation stating that the minimum age for purchasing a BB gun varies across states, whereas the reference answer is simply ``18.'' TSS generates the more direct response ``18 years,'' increasing token-level F1 from 0.043 to 0.143. The 13B QA case exhibits a larger improvement from 0.333 to 0.750. These examples indicate that TSS can reduce unnecessary elaboration and retain concise task-relevant content, which is particularly beneficial when evaluation emphasizes normalized answer overlap.

The RAG example demonstrates a partial rather than complete correction. Native generates several irrelevant filming locations and receives an F1 score of 0.000. TSS shifts the response toward the reference by mentioning locations such as Whitby, Scarborough, the City of Bradford, and the North Riding of Yorkshire, increasing F1 to 0.310. Nevertheless, the TSS output still contains additional locations and incomplete phrasing. This case shows that target skipping can preserve a better-aligned draft branch, but it does not introduce knowledge that is absent from the prompt context or candidate generation process. TSS improves the answer direction without fully resolving all factual or formatting errors.

The MMLU examples distinguish improvements in generation behavior from improvements in the final task decision. In the 7B case, both Native and TSS select option A, while the correct answer is D, so both receive an accuracy of zero. However, Native continues with an unrelated off-task passage, whereas TSS terminates with a concise answer. TSS therefore improves response stability and removes off-task drift without correcting the selected option. In the 13B case, target skipping changes an incorrect native prediction to the correct option, increasing case-level accuracy from 0 to 1. These two examples show that reducing full-depth processing may sometimes only clean the generated response, while in other cases it can also prevent later transformations from altering a correct candidate decision.

The slightly degraded 7B Translation case also highlights why TSS jointly constrains acceptance and downstream metrics during configuration search. A configuration that performs well on average may still be suboptimal for an individual request, since some prompts benefit from computations performed by the skipped layers. TSS therefore does not claim universal per-request improvement. Instead, it selects a domain-level operating point that maintains aggregate task quality while improving speculative acceptance and verification efficiency. Retaining the complete target model additionally allows unsupported domains or conservative deployments to use the original dense verification path.

Overall, the qualitative results indicate that full-depth target verification is not always optimal for domain-specific speculative decoding. TSS is particularly effective when later target transformations introduce verbosity, off-task continuation, or unnecessary revisions that reduce agreement with the task objective. At the same time, the unchanged and slightly degraded cases demonstrate that the effect is not uniform across all requests. The benefit of TSS should therefore be understood at the domain level: it improves aggregate acceptance, task performance, and efficiency, rather than assuming that every skipped layer is unnecessary for every input.

\begin{table*}[t]
\centering
\small
\caption{
Index of representative held-out qualitative cases across domains and model scales. The examples include improved, unchanged, and unsuccessful outcomes, illustrating both the benefits and limitations of domain-specific target-layer skipping.
}
\label{tab:tss_case_index}
\begin{tabular}{llcclcc}
\toprule
Model & Domain & Request ID & Skip Layers & Metric & Native & +TSS \\
\midrule
7B & Translation & \texttt{sb\_translation\_192} & $\{3,25,30\}$ & BLEU & 0.142 & 0.101 \\
7B & Summarization & \texttt{sum\_289} & $\{15,22,23\}$ & ROUGE-L & 0.281 & 0.430 \\
7B & Open-domain QA & \texttt{nq\_57} & $\{6,8,19,21,27\}$ & F1 & 0.043 & 0.143 \\
7B & RAG & \texttt{sb\_rag\_513} & $\{3,6,14,25,30\}$ & F1 & 0.000 & 0.310 \\
7B & MMLU & \texttt{mmlu\_prehistory\_53} & $\{3,7,9,14,20\}$ & Acc. & 0 & 0 \\
13B & Translation & \texttt{sb\_translation\_240} & $\{7,11,20,31,35,38\}$ & BLEU & 0.100 & 0.328 \\
13B & Summarization & \texttt{sum\_246} & $\{12,20,27,33,34,38\}$ & ROUGE-L & 0.143 & 0.280 \\
13B & Open-domain QA & \texttt{nq\_30} & $\{9,10,11,19,28,38\}$ & F1 & 0.333 & 0.750 \\

13B & MMLU & \texttt{mmlu\_professional\_law\_19} & $\{7,23,24,26,28,32,33\}$ & Acc. & 0 & 1 \\
\bottomrule
\end{tabular}
\end{table*}
\begin{table*}[t]
\centering
\small
\caption{Qualitative held-out examples for Vicuna-7B + EAGLE. Each block is one request from the official test split ($n{=}64$), comparing Native speculative decoding with the TSS skip set. Case-level task scores appear in each block header.}
\label{tab:tss_cases_7b}
\setlength{\tabcolsep}{4pt}
\begin{tabular}{@{}>{\bfseries}l p{0.90\linewidth}@{}}
\toprule
\multicolumn{2}{@{}l}{\textit{Translation} (\texttt{sb\_translation\_192}); skip $\{ 3,25,30 \}$; BLEU: 0.142$\rightarrow$\textbf{0.101}} \\
\midrule
\textbf{Input} & Je dunkler das Fleisch , desto höher der ph-Wert . \\
\textbf{Native} & The darker the meat, the higher the pH value \\
\textbf{+TSS} & The darker the meat, the higher the pH \\
\textbf{Ref.} & The darker the meat , the higher the pH value . \\
\midrule
\multicolumn{2}{@{}l}{\textit{Summarization} (\texttt{sum\_289}); skip $\{ 15,22,23 \}$; ROUGE-L: 0.281$\rightarrow$\textbf{0.430}} \\
\midrule
\textbf{Input} & Arsenal must make a swift decision on Petr Cech this summer or risk losing out on the goalkeeper. The Chelsea shot-stopper looks certain to leave Stamford Bridge at the end of the season after losing the No 1 spot to Thibaut Courtois. Cech will hold talks wit… \\
\textbf{Native} & Arsenal must make a decision on signing Petr Cech from Chelsea this summer or risk losing out on the goalkeeper. Cech is likely to leave Chelsea at the end of the season after losing his No 1 spot to… \\
\textbf{+TSS} & Arsenal must make a decision on Petr Cech this summer or risk losing out to other clubs interested in the Chelsea goalkeeper. Cech is likely to leave Stamford Bridge at the end of the season after lo… \\
\textbf{Ref.} & Arsenal risk missing out of Chelsea's Petr Cech is they hesitate on move. Cech looks set to leave Stamford Bridge after losing his No 1 spot. Liverpool, PSG, R… \\
\midrule
\multicolumn{2}{@{}l}{\textit{Open-domain QA} (\texttt{nq\_57}); skip $\{ 6,8,19,21,27 \}$; F1: 0.043$\rightarrow$\textbf{0.143}} \\
\midrule
\textbf{Input} & what age do you need to be to buy a bb gun \\
\textbf{Native} & In the United States, the minimum age to purchase a BB gun varies by state. Generally, it is legal to purchase a BB gun at the age of 18, but some states have set the minimum age at 16 or 17. It is i… \\
\textbf{+TSS} & In the United States, the minimum age to legally purchase a BB gun is 18 years. \\
\textbf{Ref.} & 18 \\
\midrule
\multicolumn{2}{@{}l}{\textit{RAG} (\texttt{sb\_rag\_513}); skip $\{ 3,6,14,25,30 \}$; F1: 0.000$\rightarrow$\textbf{0.310}} \\
\midrule
\textbf{Input} & the series is shot in Belgium, several local actors are featured: Veerle Baetens, Jurgen Delnaet, Joren Seldeslachts, Elsa Houben, Ben Forceville and Ben Van den Heuvel all appear in the series. Rebecca Ferguson who portrays Elizabeth Woodville, the White Que… \\
\textbf{Native} & The Royal was filmed in various locations in Belgium, including Bruges, Ghent, and Brussels. The Palace was filmed in Lithuania. The Crown was filmed in several locations in the UK, including London,… \\
\textbf{+TSS} & The Royal was filmed in various locations in the UK, including Whitby, Scarborough, City of Bradford, and the North Riding of Yorkshire, as well as in Lithuania. The series was also filmed in part at… \\
\textbf{Ref.} & the North Riding of Yorkshire; City of Bradford; Scarborough; Whitby \\
\midrule
\multicolumn{2}{@{}l}{\textit{MMLU} (\texttt{mmlu\_prehistory\_53}); skip $\{ 3,7,9,14,20 \}$; Acc.: 0$\rightarrow$0} \\
\midrule
\textbf{Input} & The key factor in the survival and success of anatomically modern human beings was: A. their ability to control fire and produce art. B. their ability to interbreed with other hominids. C. their ability to hunt large mammals. D. their ability to adapt. \\
\textbf{Native} & A.\ \ldots\ [off-task drift: ``2018--2023 World Outlook for ...''] \\
\textbf{+TSS} & A. their ability to control fire and produce art. \\
\textbf{Ref.} & D \\
\bottomrule
\end{tabular}
\end{table*}

\begin{table*}[t]
\centering
\small
\caption{Qualitative held-out examples for Llama-2-13B + SAMD (Token Recycle). Each block is one request from the official test split ($n{=}64$), comparing Native speculative decoding with the TSS skip set. Case-level task scores appear in each block header.}
\label{tab:tss_cases_13b}
\setlength{\tabcolsep}{4pt}
\begin{tabular}{@{}>{\bfseries}l p{0.90\linewidth}@{}}
\toprule
\multicolumn{2}{@{}l}{\textit{Translation} (\texttt{sb\_translation\_240}); skip $\{ 7,11,20,31,35,38 \}$; BLEU: 0.100$\rightarrow$\textbf{0.328}} \\
\midrule
\textbf{Input} & Ein Schuldspruch erfolgte außerdem in einer Reihe terrorismusbezogener Anklagepunkte , wozu eine Welle von Sprengstoffanschlägen auf Kirchen in Indonesien an Heiligabend im Jahr 2000 zählten . \\
\textbf{Native} & In addition to the sentence, a series of terrorism-related charges were brought, including a wave of bombings on churches in Indonesia on Christmas in the year 2000. \\
\textbf{+TSS} & A judgment was also handed down in a series of terrorism-related charges, including a wave of bombings on churches in Indonesia on Christmas Eve in the year 2000. \\
\textbf{Ref.} & He was also found guilty of a number of other terrorism-related charges , including a wave of bombings of churches across Indonesia on Christmas Eve in 2000 . \\
\midrule
\multicolumn{2}{@{}l}{\textit{Summarization} (\texttt{sum\_246}); skip $\{ 12,20,27,33,34,38 \}$; ROUGE-L: 0.143$\rightarrow$\textbf{0.280}} \\
\midrule
\textbf{Input} & Campaign: Amanda Burleigh has won her battle. A midwife has won a decade-long battle to give mothers more time attached to their newborns before the umbilical cord is cut. Amanda Burleigh was convinced that clamping the cord within seconds of delivery – shutt… \\
\textbf{Native} & Midwife Amanda Burleigh has successfully campaigned for a change in medical guidelines regarding the timing of umbilical cord clamping after birth. For decades, the practice has been to clamp the cor… \\
\textbf{+TSS} & Midwife Amanda Burleigh has successfully campaigned for a change in medical practice by convincing the National Institute for Health and Care Excellence (NICE) to change its guidelines on cord clampi… \\
\textbf{Ref.} & Amanda Burleigh was convinced clamping cord immediately was wrong. Contacted other medics and they amassed evidence it could be harmful. National Institute for… \\
\midrule
\multicolumn{2}{@{}l}{\textit{Open-domain QA} (\texttt{nq\_30}); skip $\{ 9,10,11,19,28,38 \}$; F1: 0.333$\rightarrow$\textbf{0.750}} \\
\midrule
\textbf{Input} & a request to the supreme court that it review a case that was already decided is made via a(n) \\
\textbf{Native} & Petition for Certiorari. \\
\textbf{+TSS} & Petition for a Writ of Certiorari. \\
\textbf{Ref.} & a writ of certiorari \\
\midrule
\multicolumn{2}{@{}l}{\textit{RAG} (\texttt{sb\_rag\_513}); skip $\{ 8,17,20,27,33,34 \}$; F1: 0.061$\rightarrow$\textbf{0.316}} \\
\midrule
\textbf{Input} & the series is shot in Belgium, several local actors are featured: Veerle Baetens, Jurgen Delnaet, Joren Seldeslachts, Elsa Houben, Ben Forceville and Ben Van den Heuvel all appear in the series. Rebecca Ferguson who portrays Elizabeth Woodville, the White Que… \\
\textbf{Native} & The royal TV series was filmed in various locations in the United Kingdom and Lithuania. Here are some of the specific locations where the series was shot: \\
\textbf{+TSS} & The Royal TV series was filmed in Lithuania and in the UK, specifically in Whitby, Scarborough, City of Bradford, and the North Riding of Yorkshire. The series was also filmed at Elstree Studios in B… \\
\textbf{Ref.} & the North Riding of Yorkshire; City of Bradford; Scarborough; Whitby \\
\midrule
\multicolumn{2}{@{}l}{\textit{MMLU} (\texttt{mmlu\_professional\_law\_19}); skip $\{ 7,23,24,26,28,32,33 \}$; Acc.: 0$\rightarrow$\textbf{1}} \\
\midrule
\textbf{Input} & An officer stopped a car for having a burned out headlight and license plate light. When the driver could not produce a driver's license, the officer asked him if he minded if they searched the vehicle. The officer did not advise the driver that he had a righ… \\
\textbf{Native} & C. Based on the information provided, the most likely decision of the court would be: \\
\textbf{+TSS} & A. Based on the information provided, the correct answer is: \\
\textbf{Ref.} & A \\
\bottomrule
\end{tabular}
\end{table*}
\end{document}